# DEPTH-WISE LAYERING OF 3D IMAGES USING DENSE DEPTH MAPS: A THRESHOLD BASED APPROACH

## MIRKAMALI S.S. [1*], NAGABHUSHAN P.[1]

[1]Department of Studies in Computer Science, University of Mysore, Mysore, India
*Corresponding Author: Email- s.s.mirkamali@hotmail.com

**Abstract-** Image segmentation has long been a basic problem in computer vision. Depth-wise Layering is a kind of segmentation that slices an image in a depth-wise sequence unlike the conventional image segmentation problems dealing with surface-wise decomposition. The proposed Depth-wise Layering technique uses a single depth image of a static scene to slice it into multiple layers. The technique employs a thresholding approach to segment rows of the dense depth map into smaller partitions called Line-Segments in this paper. Then, it uses the line-segment labelling method to identify number of objects and layers of the scene independently. The final stage is to link objects of the scene to their respective object-layers. We evaluate the efficiency of the proposed technique by applying that on many images along with their dense depth maps. The experiments have shown promising results of layering.

## INTRODUCTION

Image segmentation has long been a basic problem in computer vision. Depth-wise Layering is a kind of segmentation that slices an image in a depth-wise sequence unlike the conventional image segmentation problems dealing with surface-wise decomposition. In this paper, we propose a novel depth-wise image layering technique with the objective that the extracted layers not only preserve object boundaries but also maintain their depth order using information available in the depth map of the input 3D image. This technique can be used to improve viewing comfort with 3D displays [1], to compress videos [2] and to interpolate intermediate views [3]. The proposed technique, initially segments every row of the depth image using a thresholding method into smaller partitions called line-segments. Then, it uses the line-segments to extract both objects and layers of the scene independently. Finally, it links the extracted objects of layers to make object-layers. We assume that there is a depth image for each view of a static scene. The depth data could be obtained by using a range map camera, depth camera or multi-view stereo techniques [4-5]. We used both the real images and their dense depth maps in our experiments to evaluate the proposed technique. Our technique contributes the following two aspects. First, we introduce a novel thresholding base line-segmentation method which uses a deference histogram to estimate the threshold. Second, we introduce a link perception method to connect parts of the objects and make a complete object and later object-layers. The experiment results demonstrate that our Depth-wise Layering technique can reliably extract objects and their corresponding layers.

### Related Work

There are many robust image segmentation algorithms such as thresholding [6-7], mean shift [8], normalized cut [9] and watershed algorithm [10]. Mean shift based segmentation algorithms are the most reliable segmentation approaches for the task of multi-object segmentation. However, for a depth image, directly using these intensity-based segmentation methods to decompose a depth image into object-layers will result in undesirable over-segmentation because these methods are not designed to classify homogenous parts of the depth image. However, by incorporating the additional depth information and layer separation, robust segmentation can be achieved. For multi-object segmentation from the multi-view video of a dynamic scene, [11] proposed an algorithm that uses Maximum A Posterior (MAP) estimation to compute the parameters of a layered representation of the scene where each layer is modeled by its motion, appearance and occupancy. The MAP estimation of all layer parameters is equivalent to tracking multiple objects in a given number of views. Expectation-Maximization (EM) is employed to establish the layer occupancy and visibility posterior probabilities. Some other methods formulate the problem of multi-view object segmentation based on layer separation using Epipolar Plane Image (EPI) [12], a volume constructed by collecting multi-view images taken from equidistant locations along a line. This category of methods suffer from an important limitation: all of them classify moving objects as the first layer of a scene and other objects are left as the background layer. Our work is closely related to depth-based segmentation proposed by [2]. They used a Markovian statistical approach to segment the depth map into regions homogeneous in depth. In their approach two types of information are used to overcome the segmentation, which are depth map and stereoscopic prediction error. Another approach, proposed by [3], also worked on layered representation of scenes based on multiple image analysis. In their approach, the problem of segmentation is formulated to exploitation of the color and edge information. The color information is analyzed separately in YUV-channels and edge information is used in intermediate processes. The proposed method works fine in many cases. However, in case of periodic texture in the background both the disparity and color segmentation processes will fail.

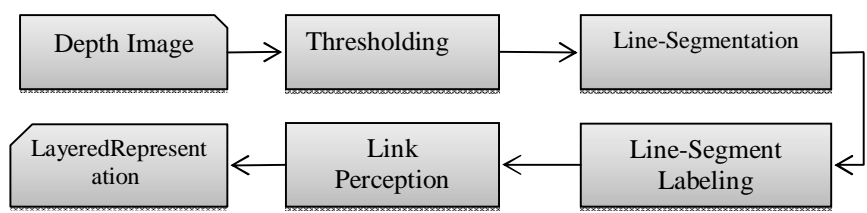

Fig.1- Block diagram of the proposed Depth-wise Layeringtechnique.

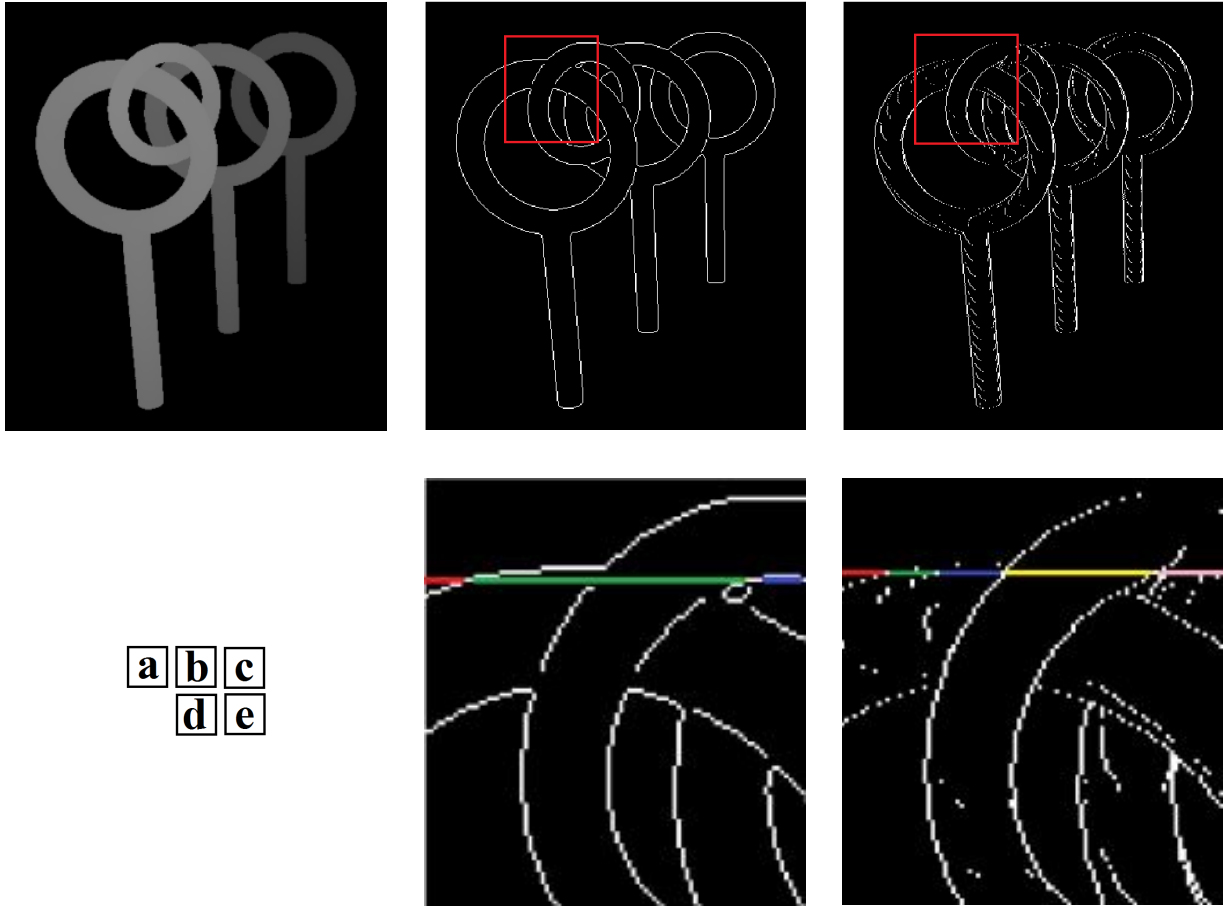

Fig.2- (a) Rings depth image (b) result of using Canny edge detection for line segmentation; (c) result of our line segmention approach; (d-e) magnified portion of (b) and (c) respectively (each color in these images, showes a separate line-segment).

### THE PROPOSED DEPTH-WISE LAYERING TECHNIQUE

Suppose that a 3D image $I$ having a depth image $D$ with $L$ gray levels $L = \{0, 1, \ldots, L-1\}$, containing the $Z$ value of an image $I$, is to be sliced into $m$ object-layers $OL = \{ol_1, ol_2, \ldots, Ol_m\}$.

The proposed segmentation technique is to first decompose the objects into some line-segments using a thresholding method and then to link the line-segments of an object to make compound objects and later object-layers. The major steps of this method are summarized in Fig. 1.

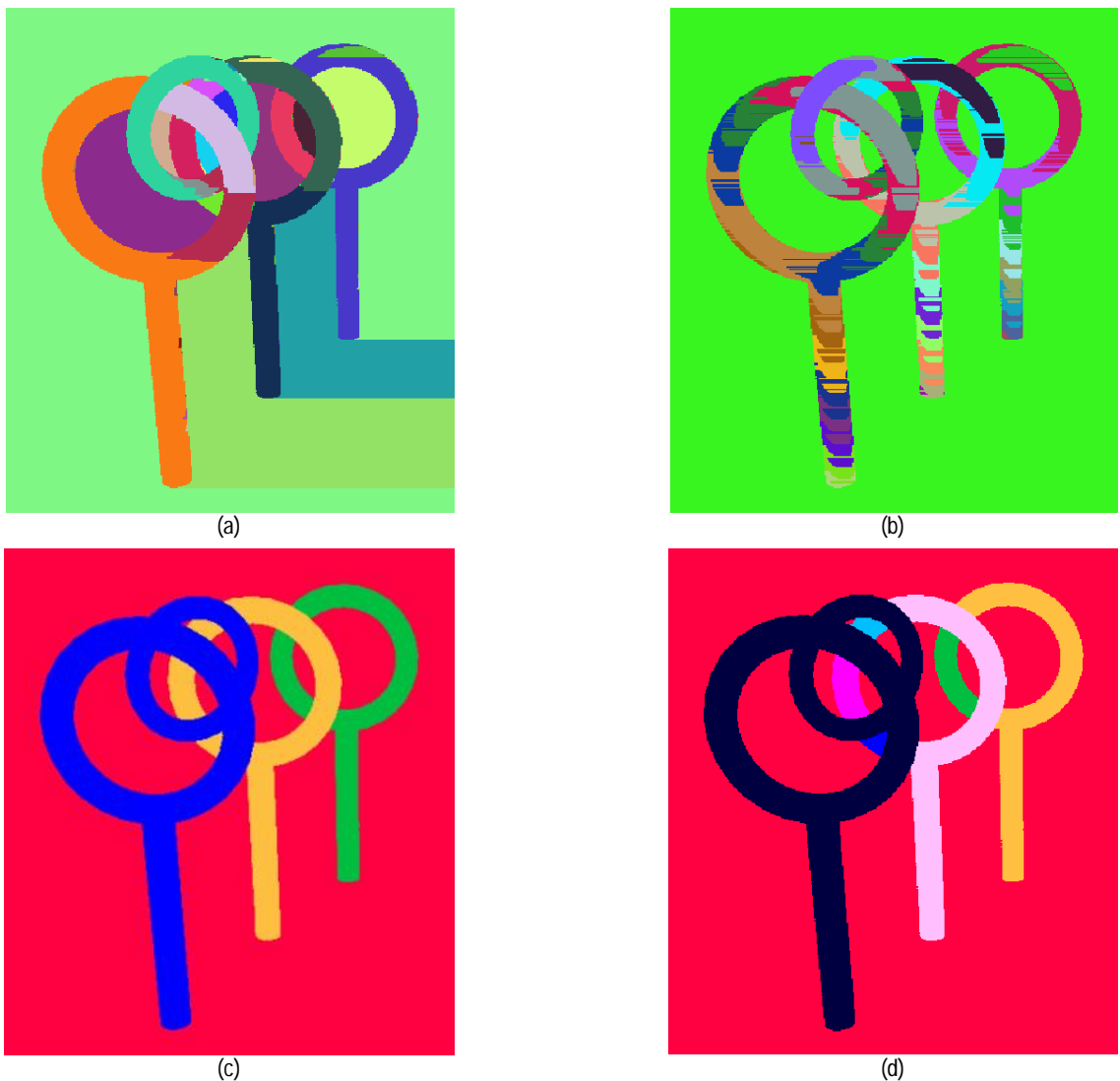

Fig.3- Intermediate results of the: (a) line-segment labelling based on object numbers; (b) line-segment labelling based on layer numbers; (c) combined segmentation result after applying the link perception method; (d) Mean-Sift segmentation result.

**Line Segmentation with Thresholding**

A sharp change of gray value in a row of a depth image shows a change between two objects in two different layers. Finding the sharp changes is a fundamental problem of edge detection techniques. However, directly using an edge detection method to find the line-segments is not giving a proper result. Because edges extracted from depth images using general purpose edge detection methods often hampered by fragmentation, missing edge segments, as well as miss positioned edges especially in ramp and steps. An example of edge map of depth image obtained using canny edge detection is shown in fig.2-b. As it is shown in fig.2-b and fig.2-d, using canny edge detection will result in incorrect line segmentation. This failure is because of the fragmented edges of the objects. Our line-segmentation method takes as input the depth image and produces line-segments for every row of a depth image $D$. The line-segmentation pipeline used in our paper consists of a threshold estimation process followed by an edge localization process.

Edge localization is a process of finding start and end points of line-segments in a depth image. In this process we are dealing with local computation. The method of choice to determine whether a depth value is a starting point of a line-segment $LS$ or not, is to use a threshold. A technique for automatic calculation of the threshold values $T_r$ and $T_c$ is calculating the $max(dh(n_k))$ for a random number of the rows and columns respectively. The threshold estimator of a depth image with gray levels in the range $[0, L-1]$ is a discrete function $dh(n_k) = n_k$, where $n_k$ the difference between two consecutive pixels and $n_k$ is the number of pixels in $k^{th}$ row/column having difference value of $n_k$. We define a point as being a starting point of a line-segment if its deference value is greater than a threshold $T_r$. The threshold value $T_c$ is defined as a connectivity value of objects which is described in the next section.

Advantage of using the proposed edge localization method is in extraction of all the vertically connected edges (Fig .2-c and 2.e) which is significantly important for the proposed line-segment labelling algorithm.

**Line-Segment Labeling**

In the proposed Line-Segment Labeling technique, we assigned two attributes to a line-segment $LS$ as its representatives. The first is object number which controls the connection between line-segments vertically. We define an object-number (ON) of a line-segment as

$$LS_{i,j}.ON = \begin{cases} LS_{i-1,k}.ON & if LS_{i,j}.ON \cap LS_{i-1,k}.ON \neq \emptyset \\ & and\ |V_{i,j} - V_{i-1,k}| \leq T\ k = 1 \ldots p \\ oc + 1 & otherwise \end{cases} \quad (5)$$

where $i$ is the row number of the depth map $D$ in which a line-segment belongs, $j$ is a number of a line-segment in $i^{th}$ row, $k$ is a number of a line-segment in $(i-1)^{th}$ row, $p$ is the number of line-segments in $(i-1)^{th}$ row, $T$ is object connectivity value which is calculated manually, $oc$ is the number of available object values and

$$V_{i,j} = mode(D(i,q))\ q = s_{i,j}\ to\ e_{i,j} \quad (6)$$

where $s$ and $e$ are starting and ending points of a line-segment respectively. An example of extracted object numbers is shown in Fig.3-a. In this image each color shows a different object number.

The second feature of a line-segment is its layer-number. A layer-number (LN) of a line-segment is defined as

$$LS_{i,j}.LN = \begin{cases} k\ if\ V_{i,j} = V_k\ for\ k = 1 \ldots lc \\ n+1 & otherwise \end{cases} \quad (7)$$

where $lc$ is denoting the number of available layers and $V_k$ is the depth value of the $k^{th}$ layer ($L_k$). An example of extracted layer numbers is shown in Fig .3-b. In this image each color shows a different layer number.

**Link Perception**

The idea behind the proposed link perception method is based on two assumptions. First, every object should appear only in one object-layer completely. Second, all the objects in a layer are in one object-layer.

The first step in this algorithm is to link all the parts of an object and all the layers to which the object belongs and make a compound object $CO$ using condition (8).

$$\begin{cases} CO_l.LN = CO_l.LN \cup LS_{i,j}.LN \\ \quad if LS_{i,j}.ON = CO_l.ON \\ CO_{l+1}.LN = LS_{i,j}.LN \\ CO_{l+1}.ON = LS_{i,j}.ON \end{cases} otherwise \quad (8)$$

where $CO.LN$ and $CO.ON$ denote the layer number and object number of a compound object $CO$ respectively.

The second step is to link all the compound objects which are in the same layer to construct an object-layer using the following condition.

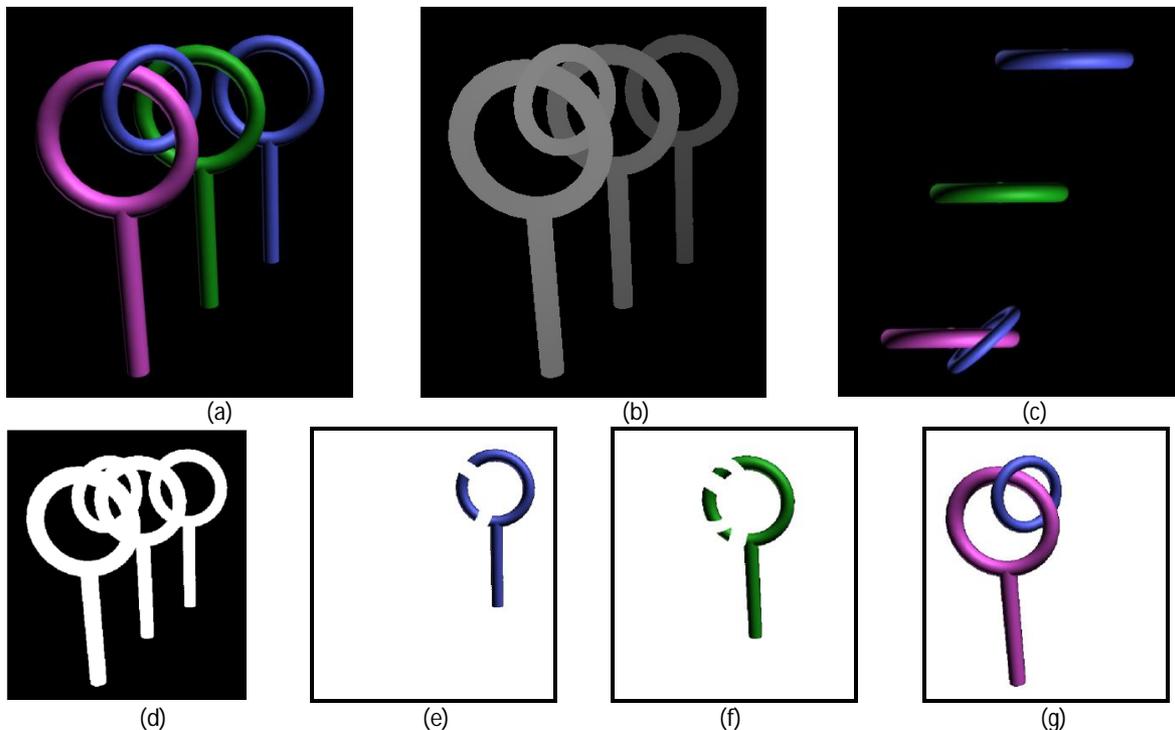

Fig.4- Layered representation of a computer generated scene; (a) original image; (b) depth map; (c) top view of the scene; (d) layered representation of the background; (e) first layer; (f) second layer; (g) third layer.

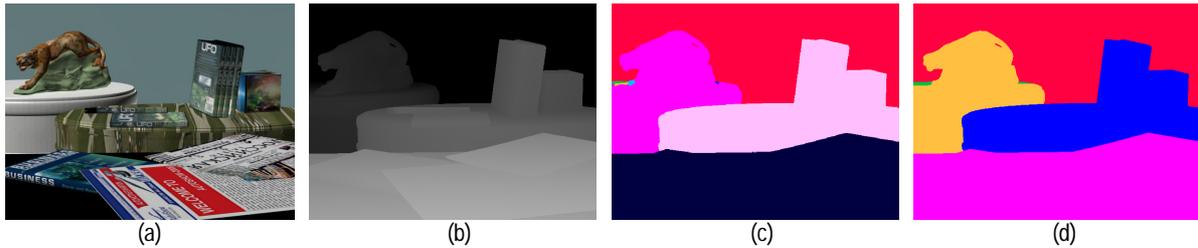

Fig.5- Another example of a computer generated scene (a) original image; (b) depth map; (c) color coded object-layers obtained from our segmentation technique; (d) object-layers obtained from Mean Shift segmentation technique.

$$OL_m = OL_m \cup CO_l \text{ if } CO_l.LN \cap OL_m.LN \neq \emptyset \quad (9)$$

where $l$ is the compound object number and m is an object-layer number. An example of extracted object-layers is shown in Fig.3-c.

**EXPERIMENTAL RESULTS**

To verify the efficiency of the technique, experiments have been carried on many images along with their depth map. Since a dataset containing clutter of linearly separable objects of the captured static scenes, with their proper dense depth map yet does not exist in community, we chose to use computer generated 3D images and their dense depth map.

One of the challenges of image segmentation techniques is over-segmentation of occluded objects. In these cases, an object will separate into some parts because of lack of knowledge about the parts of the objects. Fig.3 shows one of those challenging examples which three rings are positioned in a line toward the camera and occluded by their neighbouring objects. Fig.3-c shows the result of our proposed layering algorithm which preserves all the parts of a ring in a class. On the other hand, Fig. 3-d shows the result of the mean-shift segmentation applied on the same depth image that is Fig. 4-b. It is clearly visible that each ring in the second and third positions is separated into a few different classes; each class is shown by a different color.

In this paper, we used two critical images to illustrate the efficiency of the proposed technique. Figs. 4 and 5 show the examples as well as the results with (a) showing the original 3D image, (b) showing depth image (darker colors representing larger z values), (c) showing the top down view of the 3D scene to compare the extracted layers visually.

Fig.4-a shows an image containing three rings. This image has been created to show how robust the proposed technique is against the occlusion. The depth map and top view of the scene are given in Fig.4-b and Fig.4-c. The layered representation of image in Fig. 4(a) is shown in Figs. 4-d to (g). The four layers correspond to (a) background (e-g) three rings each in a different object-layer. It is clearly visible that the main features of all the three rings are well preserved and all of them are well classified into their corresponding layers using the proposed technique.

Fig.5-a is an image of a clutter of office objects in which some of them are occluded and positioned in different layers. Fig.5-b is the depth map of the given 3D image, Fig.5-c color coded object-layers obtained from our segmentation technique and Fig.5-d object-layers obtained from Mean Shift segmentation technique.

**CONCLUSION**

In the paper, we proposed an image segmentation technique called Depth-wise Layering. The technique is capable of layering an image into multiple layers based on the position of objects in the scene with respect to the camera. A threshold based algorithm was used to divide a depth map into line-segments. After assigning an object number and a layer number to every line-segment, we employed a link perception algorithm to compose divided objects and make the completed objects. Later, all the objects in a layer were linked to make object-layers. We conducted experiments on challenging examples in which the scene contained occluded objects. Results showed that our proposed technique gave good performance.